\documentclass[10pt,twocolumn,letterpaper]{article}

\usepackage[pagenumbers]{iccv} %

\usepackage{paralist}
\usepackage{enumitem}
\usepackage{multirow}
\usepackage{titletoc}
\usepackage{booktabs}
\usepackage{gensymb}

\newcommand{\vect}[1]{\boldsymbol{\mathbf{#1}}}

\newcommand{\methodname}{DNF-Avatar\xspace}
\newcommand{\suppmat}{Supp. Mat\xspace}

\newcommand{\figref}[1]{Fig.~\ref{#1}}
\newcommand{\tabref}[1]{Tab.~\ref{#1}}
\newcommand{\secref}[1]{Sec.~\ref{#1}}

\newcommand{\equref}[1]{Equ.~(\ref{#1})}

\definecolor{iccvblue}{rgb}{0.21,0.49,0.74}
\definecolor{myPink}{RGB}{255,105,180}
\usepackage[pagebackref,breaklinks,colorlinks,allcolors=iccvblue,urlcolor=myPink]{hyperref}

\def\paperID{1777} %
\def\confName{ICCV}
\def\confYear{2025}

\newcommand{\boldparagraph}[1]{\vspace{0.0cm}\noindent{\bf #1:} }

\title{DNF-Avatar: Distilling Neural Fields for Real-time Animatable Avatar Relighting}

\newcommand\rurl[1]{%
  \href{https://#1}{\nolinkurl{#1}}%
}

\author{Zeren Jiang$^{1}$
\quad
Shaofei Wang$^{2}$
\quad
Siyu Tang$^{2}$ \\
 $^1$Visual Geometry Group, University of Oxford \quad 
 $^2$ETH Zürich \\
{\tt\small zeren@robots.ox.ac.uk \quad \{shaofei.wang, siyu.tang\}@inf.ethz.ch}\\[0.1em]
\small\rurl{jzr99.github.io/DNF-Avatar}
}

\begin{document}
\twocolumn[{%
\renewcommand\twocolumn[1][]{#1}%
\maketitle
\vspace{-4.0em}
\begin{center}
    \captionsetup{type=figure}
   \includegraphics[width=\linewidth,trim=0 5 0 0,clip]{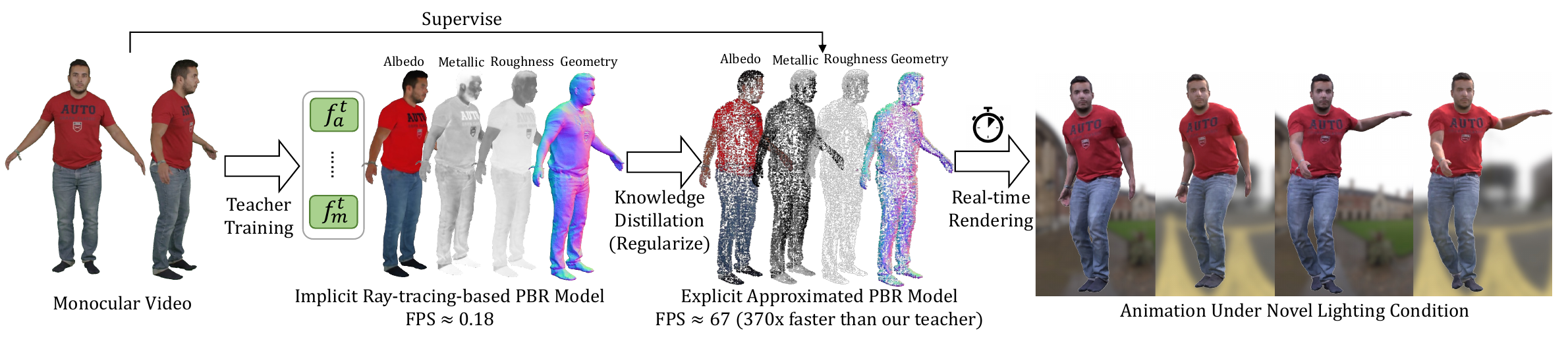}
    \vspace{-0.7cm}
    \captionof{figure}{We propose \methodname, which distills an implicit
    ray-tracing-based relightable avatar model into a Gaussian-splatting-based
    representation for real-time rendering and relighting. Our method achieves
    relighting results that are comparable to the teacher model while being 370
    times faster at inference time, achieving a 67 FPS rendering speed under
    unseen environment lighting and unseen poses.}
    \label{fig:teaser}
\end{center}%

}]

\newcommand{\tableRelighting}{
\begin{table}[ht]
\vspace{-0.3cm}
\centering
\begin{tabular}{@{}l|c|ccc@{}}
\toprule
\multirow{2}*{Method} & \multirow{2}*{\textbf{FPS}} & \multicolumn{3}{c}{\textbf{Relighting}} \\ 
~ & ~ & \textbf{PSNR}~$\uparrow$ & \textbf{SSIM}~$\uparrow$ & \textbf{LPIPS}~$\downarrow$  \\ 
\midrule
R4D~\cite{Chen2022ECCVa} & 0.25 & 16.62 & 0.8370 & 0.1726 \\

IA~\cite{Wang2024CVPR}  & 0.18 & 18.18 & 0.8722 & \underline{0.1279} \\
 \midrule
 Ours-D  & \underline{33} & \underline{18.93} & \underline{0.8768} & \textbf{0.1275} \\
 Ours-F & \textbf{67} & \textbf{19.04} & \textbf{0.8772} & 0.1307 \\
\bottomrule
\end{tabular}
\vspace{-0.3cm}
\caption{\textbf{Quantitative Relighting Results on RANA.} Our model achieves comparable or even superior results (on PSNR and SSIM) with teacher model IA~\cite{Wang2024CVPR}, while being hundreds of times faster.}
\label{tab:relighting}
\vspace{-0.3cm}
\end{table}
}

\newcommand{\tableProperty}{
\begin{table}[ht]
\vspace{-0.00cm}
\centering
\resizebox{\linewidth}{!}{%
\begin{tabular}{@{}l|c|ccc@{}}
\toprule
\multirow{2}*{Method} & \multirow{2}*{\textbf{Normal}~$\downarrow$} & \multicolumn{3}{c}{\textbf{Albedo}} \\ 
~ & ~ & \textbf{PSNR}~$\uparrow$ & \textbf{SSIM}~$\uparrow$ & \textbf{LPIPS}~$\downarrow$  \\ 
\midrule
R4D~\cite{Chen2022ECCVa} & 27.38 $^{\circ}$ & 18.23 & 0.8254 & 0.2043 \\

IA~\cite{Wang2024CVPR}  & 9.96 $^{\circ}$ & \textbf{22.83} & \textbf{0.8816} & 0.1617 \\
 \midrule
 Ours  & \textbf{9.58} $^{\circ}$ & 22.65 & 0.8701 & \textbf{0.1561} \\
\bottomrule
\end{tabular}
}
\vspace{-0.3cm}
\caption{\textbf{Quantitative Decomposition Results on RANA.} Our model surpasses the teacher in terms of normal accuracy and achieves comparable results in albedo estimation.}
\label{tab:decomposition}
\vspace{-0.7cm}
\end{table}
}

\newcommand{\tableAblation}{
\begin{table}[ht]
\vspace{-0.10cm}
\centering
\resizebox{\linewidth}{!}{%
\begin{tabular}{@{}l|ccc@{}}
\toprule
\multirow{2}*{Method} & \multicolumn{3}{c}{\textbf{Relighting}} \\ 
~  & \textbf{PSNR}~$\uparrow$ & \textbf{SSIM}~$\uparrow$ & \textbf{LPIPS}~$\downarrow$  \\ 
\midrule
w/o distillation & 18.99 & 0.8739 & 0.1488 \\

w/o distillation avatar   & 19.47 & 0.8878 & 0.1332 \\

w/o part-wise occ.  & 19.43 & 0.8854 & 0.1344 \\
 \midrule
 Ours  & \textbf{19.48} & \textbf{0.8884} & \textbf{0.1315} \\
\bottomrule
\end{tabular}
}
\vspace{-0.3cm}
\caption{\textbf{Quantitative Ablation Studies on RANA.} All components effectively contribute to the final relighting quality. }
\label{tab:ablation}
\vspace{-0.60cm}
\end{table}
}

\newcommand{\tableFPS}{
\begin{table}[ht]
\vspace{-0.1cm}
\centering
\begin{tabular}{l|cccc|c}
\toprule
Method & \textbf{LBS} & \textbf{Occ.} & \textbf{Shading} & \textbf{Rast.} & \textbf{Total} \\ 
\midrule
Ours-D & 3.3ms & 7.7ms & 12.1ms & 6.9ms & 30.0ms \\

Ours-F  & 3.3ms & 7.7ms & 0.9ms & 2.9ms & 14.8ms \\

\bottomrule
\end{tabular}
\vspace{-0.3cm}
\caption{\textbf{Time cost for each part of our model.} }
\label{tab:fps}
\vspace{-0.5cm}
\end{table}
}

\newcommand{\tablesuppdistill}{
\begin{table}[ht]
\vspace{-0.0cm}
\centering
\resizebox{\linewidth}{!}{%
\begin{tabular}{@{}l|c|ccc@{}}
\toprule
\multirow{2}*{Method} & \multirow{2}*{\textbf{Normal}~$\downarrow$} & \multicolumn{3}{c}{\textbf{Relighting}} \\ 
~ & ~ & \textbf{PSNR}~$\uparrow$ & \textbf{SSIM}~$\uparrow$ & \textbf{LPIPS}~$\downarrow$  \\ 
\midrule
R4D~\cite{Chen2022ECCVa} & 33.61 \degree & 18.22 & 0.8425 & 0.1612 \\
IA~\cite{Wang2024CVPR} & 12.05 \degree & 18.48 & 0.8859 & \textbf{0.1219} \\
\midrule
w/o dist.  & 16.49 \degree & 18.99 & 0.8739 & 0.1488 \\
w/o i-dist.  & 14.55 \degree & 19.30 & \textbf{0.8889} & 0.1392 \\
w/o p-dist. & 11.56 \degree & 19.42 & 0.8835 & 0.1374 \\
w/o dist. avatar & \underline{11.50} \degree & \underline{19.47} & 0.8878 & \underline{0.1332} \\
Ours & \textbf{11.41} \degree & \textbf{19.48} & \underline{0.8884} & \textbf{0.1315} \\
\bottomrule
\end{tabular}
}
\vspace{-0.3cm}
\caption{\textbf{Quantitative Ablation Studies on RANA.} Both objectives for distillation effectively contribute to the final relighting quality. }
\label{tab:supp_distill}
\vspace{-0.3cm}
\end{table}
}

\newcommand{\tablesupprana}{
\begin{table*}[t]
    \centering
    \vspace{1.5cm}
    \begin{tabular}{l|c|c|c|c|c|c|c|c}
        \toprule
        \multirow{2}{*}{Subject} & \multirow{2}{*}{Method} & \multicolumn{3}{c|}{Albedo} & \multirow{1}{*}{Normal} & \multicolumn{3}{c}{Relighting (Novel Pose)} \\
        \cmidrule{3-5} \cmidrule{7-9}
        & & PSNR $\uparrow$ & SSIM $\uparrow$ & LPIPS $\downarrow$ & \multicolumn{1}{c|}{Error $\downarrow$} & PSNR $\uparrow$ & SSIM $\uparrow$ & LPIPS $\downarrow$ \\
        \cmidrule{1-9}
        \multirow{4}{*}{Subject 01}
        & R4D & 20.04 & 0.8525 & 0.2079 & 33.61 \degree & 18.22 & 0.8425 & 0.1612 \\
        & IA & {24.11} & {0.8679} & {0.1827} & 12.05 \degree & {18.48} & {0.8859} & 0.1219 \\
        & Ours-D & \multirow{2}{*}{23.90} & \multirow{2}{*}{0.8580} & \multirow{2}{*}{0.1834} & \multirow{2}{*}{11.41} \degree & 19.42 & 0.8905 & 0.1252  \\
        & Ours-F & ~ & ~ & ~ & ~ & 19.48 & 0.8884 & 0.1315 \\
        \cmidrule{1-9}
        \multirow{4}{*}{Subject 02}
        & R4D & 12.13 & 0.7690 & 0.2599 & 28.34 \degree & 14.38 & 0.8128 & 0.1787 \\
        & IA & {20.94} & {0.8892} & {0.1854} & 9.29 \degree & {19.08} & {0.8812} & {0.1323} \\
        & Ours-D & \multirow{2}{*}{20.76} & \multirow{2}{*}{0.8773} & \multirow{2}{*}{0.1675} & \multirow{2}{*}{9.04} \degree & 19.86 & 0.8875 & 0.1285 \\
        & Ours-F & ~ & ~ & ~ & ~ & 20.03 & 0.8891 & 0.1297 \\
        \cmidrule{1-9}
        \multirow{4}{*}{Subject 05}
        & R4D & 19.74 & 0.8151 & 0.2488 & 26.14 \degree & {17.72} & 0.8469 & 0.1780 \\
        & IA & {22.24} & {0.8591} & {0.2071} & 9.52 \degree & 17.47 & {0.8769} & {0.1453} \\
        & Ours-D & \multirow{2}{*}{22.26} & \multirow{2}{*}{0.8527} & \multirow{2}{*}{0.1798} & \multirow{2}{*}{9.07} \degree & 18.89 & 0.8876 & 0.1377 \\
        & Ours-F & ~ & ~ & ~ & ~ & 18.97 & 0.8873 & 0.1411 \\
        \cmidrule{1-9}
        \multirow{4}{*}{Subject 06}
        & R4D & 21.57 & 0.7992 & 0.2177 & 25.83 \degree & 17.54 & 0.8866 & 0.1636 \\
        & IA & {22.94} & {0.8233} & {0.1928} & {8.89} \degree & {18.14} & {0.8932} & {0.1271} \\
        & Ours-D & \multirow{2}{*}{22.91} & \multirow{2}{*}{0.8163} & \multirow{2}{*}{0.1752} & \multirow{2}{*}{9.03} \degree & 18.67 & 0.8960 & 0.1289 \\
        & Ours-F & ~ & ~ & ~ & ~ & 18.72 & 0.8953 & 0.1341 \\
        \cmidrule{1-9}
        \multirow{4}{*}{Subject 33}
        & R4D & 18.35 & 0.8426 & 0.1887 & 25.24 \degree & 16.78 & 0.8173 & 0.1859 \\
        & IA & 21.67 & {0.8703} & 0.1351 & 9.52 \degree & {18.03} & {0.8426} & 0.1366 \\
        & Ours-D & \multirow{2}{*}{21.18} & \multirow{2}{*}{0.8450} & \multirow{2}{*}{0.1544} & \multirow{2}{*}{8.92} \degree & 19.13 & 0.8546 & 0.1331 \\
        & Ours-F & ~ & ~ & ~ & ~ & 19.23 & 0.8557 & 0.1332\\
        \cmidrule{1-9}
        \multirow{4}{*}{Subject 36}
        & R4D & 23.80 & {0.9100} & 0.1611 & 24.76 \degree & 17.05 & 0.8574 & 0.1707 \\
        & IA & {24.88} & 0.8900 & {0.1324} & 9.22 \degree & {17.46} & {0.8726} & {0.1284} \\
        & Ours-D & \multirow{2}{*}{24.43} & \multirow{2}{*}{0.8785} & \multirow{2}{*}{0.1384} & \multirow{2}{*}{9.27} \degree & 18.18 & 0.8764 & 0.1293 \\
        & Ours-F & ~ & ~ & ~ & ~ & 18.26 & 0.8773 & 0.1389 \\
        \cmidrule{1-9}
        \multirow{4}{*}{Subject 46}
        & R4D & 18.13 & 0.8777 & 0.1238 & 33.27 \degree & 16.30 & 0.8338 & 0.1649 \\
        & IA & {22.47} & {0.9391} & {0.0725} & 10.69 \degree & {17.08} & {0.8406} & 0.1000 \\
        & Ours-D & \multirow{2}{*}{22.36} & \multirow{2}{*}{0.9298} & \multirow{2}{*}{0.0793} & \multirow{2}{*}{10.25} \degree & 17.47 & 0.8415 & 0.1039 \\
        & Ours-F & ~ & ~ & ~ & ~ & 17.62 & 0.8426 & 0.1041 \\
        \cmidrule{1-9}
        \multirow{4}{*}{Subject 48}
        & R4D & 12.10 & 0.7370 & 0.2264 & 21.84 \degree & 14.98 & 0.7985 & 0.1776 \\
        & IA & {23.36} & {0.9137} & 0.1857 & 10.49 \degree & {19.70} & {0.8849} & 0.1313 \\
        & Ours-D & \multirow{2}{*}{23.39} & \multirow{2}{*}{0.9034} & \multirow{2}{*}{0.1707} & \multirow{2}{*}{9.62} \degree & 19.82 & 0.8808 & 0.1329 \\
        & Ours-F & ~ & ~ & ~ & ~ & 19.97 & 0.8816 & 0.1328 \\
        \cmidrule{1-9}
        \multirow{4}{*}{Average}
        & R4D* & 18.23 & 0.8254 & 0.2043 & 27.38 \degree & 16.62 & 0.8370 & 0.1726 \\
        & IA & \textbf{22.83} & \textbf{0.8816} & {0.1617} & 9.96 \degree & {18.18} & {0.8722} & 0.1279 \\
        & Ours-D & \multirow{2}{*}{22.65} & \multirow{2}{*}{0.8701} & \multirow{2}{*}{\textbf{0.1561}} & \multirow{2}{*}{\textbf{9.58}} \degree & 18.93 & 0.8769 & \textbf{0.1275} \\
        & Ours-F & ~ & ~ & ~ & ~ & \textbf{19.04} & \textbf{0.8772} & 0.1307 \\
        \bottomrule
    \end{tabular}
\caption{\textbf{Per-Subject Metrics on the RANA dataset.}}
    \label{tab:metric_all_rana}
    \vspace{1.5cm}
\end{table*}
}

\begin{abstract}
Creating relightable and animatable human avatars from monocular videos is a rising research topic with a range of applications, \eg\ virtual reality, sports, and video games. Previous works utilize neural fields together with physically based rendering (PBR), to estimate geometry and disentangle appearance properties of human avatars. However, one drawback of these methods is the slow rendering speed due to the expensive Monte Carlo ray tracing.  To tackle this problem, we proposed to distill the knowledge from implicit neural fields (teacher) to explicit 2D Gaussian splatting (student) representation to take advantage of the fast rasterization property of Gaussian splatting.  To avoid ray-tracing, we employ the split-sum approximation for PBR appearance.  We also propose novel part-wise ambient occlusion probes for shadow computation.  Shadow prediction is achieved by querying these probes only once per pixel, which paves the way for real-time relighting of avatars.  These techniques combined give high-quality relighting results with realistic shadow effects.  Our experiments demonstrate that the proposed student model achieves comparable or even better relighting results with our teacher model while being 370 times faster at inference time, achieving a 67 FPS rendering speed.
\end{abstract}

\vspace{-1.2em}
\section{Introduction}
\label{sec:intro}
Reconstructing animatable human avatars with relightable appearance is an
emerging research topic in computer vision and computer graphics. It has a wide
range of applications, such as virtual reality, sports, and video games.
Traditional
methods~\cite{Li2013Eurographics,Dou2016SIGGRAPH,Collet2015SIGGRAPH,Prada2017SIGGRAPH,Guo2019SIGGRAPH,Edoardo2022SIGGRAPH,Isik2023SIGGRAPH}
for creating human avatars require dense multi-view capturing systems, which are
expensive and not scalable.  To enable a relightable appearance, controlled lighting
conditions are also required, which further complicates the capturing
process~\cite{Debevec2000SIGGRAPH,Theobalt2007TVCG,Iwase2023CVPR,Chen2024CVPR,Saito2024CVPR}.
Overall, these traditional methods are inaccessible to the general public due
to their high cost and complexity.

In recent years, researchers have proposed methods to create animatable~\cite{Chen2023CVPR,Jiang2024CVPR, reloo, xue2024hsr} and relightable~\cite{Wang2024CVPR, Xu2024CVPR, Lin2024AAAI, Chen2022ECCVa} human
avatars using neural fields~\cite{Park2019CVPR,Mescheder2019CVPR} along with
human body prior models~\cite{Loper2015SIGGRAPHASIA}.  The robustness of neural fields allows for the
estimation of geometry and appearance properties from monocular videos.
However, one drawback of these methods is the slow rendering speed due to the
underlying neural radiance fields (NeRF~\cite{Mildenhall2020ECCV})
representation and the use of physically based rendering (PBR).
To achieve PBR, existing methods employ Monte Carlo ray tracing, which is accurate but usually requires tracing a large number of secondary rays to attain
high-quality PBR results, whereas a typical NeRF model only requires tracing a
single primary ray to render a pixel.  Thus, even with various acceleration
techniques for NeRF, the rendering of state-of-the-art relightable human
avatars is still inefficient, taking several seconds to render a single frame \cite{Wang2024CVPR, Xu2024CVPR, Lin2024AAAI, Chen2022ECCVa}.

With the advent of 3D Gaussian splatting (3DGS~\cite{Kerbl2023TOG} and follow-up
2DGS~\cite{Huang2024SIGGRAPH}), a bunch of works has shown that Gaussian
splatting can achieve real-time rendering of human avatars when combined with
human body prior
models~\cite{Qian2024CVPR,Moreau2024CVPR,Pang2024CVPR,Li2024CVPR,Kocabas2024CVPR,Lei2024CVPR,Hu2024CVPR,Jena2023ARXIV,Zielonka2023ARXIV,Ye2023ARXIV,Liu2023ARXIV,guo2025vid2avatarpro,zheng2025gaustar,zhang2025odhsronlinedense3d}.
However, the majority of these works focus on the novel-view synthesis task and
do not consider the relightable appearance.

There are two major challenges in extending Gaussian splatting to relightable
human avatars: (1) The vanilla 3DGS does not produce high-quality geometric
details compared to NeRF-based
methods~\cite{Oechsle2021ICCV,Yariv2021NEURIPS,Wang2021NEURIPSa}, which is
crucial for relighting.  (2) Monte Carlo estimation of PBR incurs a significant
computational overhead, which nullifies the advantage of real-time rendering from
Gaussian splatting techniques.  Recent methods~\cite{Lin2024AAAI,Zhao2024ARXIV} avoid
expensive ray-tracing by using efficient pre-trained/cached visibility models.  However,
they still require querying the visibility models multiple times per pixel, preventing real-time
performance.

To address the first challenge, we use the recently proposed 2D Gaussian
splatting (2DGS~\cite{Huang2024SIGGRAPH}) representation, which can achieve
improved geometry reconstruction compared to vanilla 3DGS.  We note that Gaussian-splatting-based
methods are less robust than NeRF-based methods during training, especially when
the number of input views is limited.
We thus propose to distill the normal prediction from
a pre-trained neural-field-based teacher model~\cite{Wang2024CVPR} to an explicit 2DGS-based
student model to achieve high-quality geometry reconstruction.  To address the
second challenge, we use a split-sum approximation for the specular
appearance.  We also introduce novel part-wise ambient occlusion probes to enable
efficient shadow computation of articulated bodies; it achieves shadow prediction 
with \textit{a single query} to the probes, which is crucial for our final real-time
rendering performance.
Lastly, the split-sum approximation is
less physically plausible compared to ray-tracing-based PBR; thus, we utilize
the ray-tracing-based teacher model~\cite{Wang2024CVPR} to further regularize the
student model's material prediction during training.  These techniques combined
allow us to achieve high-quality relighting results with realistic shadow
effects, while circumventing the time-consuming ray tracing in PBR, thereby
enabling real-time relighting ($67$ FPS) under arbitrary novel poses.

In summary, our contributions are:
\begin{compactitem}
 \item A novel framework that creates animatable and relightable avatars for
     real-time rendering based on an approximated PBR pipeline.
 \item Knowledge distillation strategy between implicit neural field and
     explicit 2DGS representation for human avatar reconstruction.
 \item Novel precomputed part-wise ambient occlusion probes that lead to fast and
     high-fidelity shadow modeling.
\end{compactitem}

\section{Related Work}
\label{sec:related}
\subsection{Radiance Field Representations}
Since the emergence of Neural Radiance Fields (NeRF)~\cite{Mildenhall2020ECCV},
many follow-up works have been proposed to improve different aspects of NeRF.
~\cite{Yariv2021NEURIPS,Wang2021NEURIPSa} and~\cite{Oechsle2021ICCV} have
proposed to use the signed distance field and occupancy field, respectively, to
replace the density field used in the vanilla NeRF.  This achieves improved
geometry reconstruction quality.  Another line of
works have focused on accelerating the training and inference speed of NeRF,
they combine NeRF with various accelerating data structures, including hash
grid~\cite{Mueller2022SIGGRAPH}, tri-planes~\cite{Chan2022CVPR},
voxels~\cite{AlexYuandSaraFridovich-Keil2022CVPR,Sun2022CVPR}, and
tensor-decomposition~\cite{Chen2022ECCV} to achieve fast training and inference
of NeRFs.  Last but not
least,~\cite{Zhang2021SIGGRAPHASIA,Jin2023CVPR,Zhang2022CVPRa} proposed to use
neural fields to represent intrinsic properties, such as albedo and roughness, to
enable scene relighting.

Contrary to NeRF representations that use implicit neural fields to predict
properties of arbitrary points in 3D space, point-based explicit
representations~\cite{Ruckert2022TOG, Xu2022CVPRa,Zhang2022ARXIV}
instead store rendering-related properties in point-based primitives. This kind
of representation enables fast rasterization~\cite{Schutz2022CGIT} and is
efficient and flexible to represent intricate structures. Notably,
3DGS~\cite{Kerbl2023TOG} leverage 3D Gaussian as primitives to represent
radiance field, achieving state-of-the-art rendering quality with real-time
inference speed.  2DGS~\cite{Huang2024SIGGRAPH} further enhanced 3DGS by replacing
3D Gaussians with 2D Gaussians to enable
multi-view consistent rendering of Gaussian primitives, thus achieving
high-quality geometry reconstruction. However, compared to neural field
representations which is robust even under sparse input views, the explicit
point-based representation often requires dense input views with good
initialization and regularization to achieve high-quality results.

\subsection{Knowledge Distillation}
Knowledge distillation~\cite{Hinton2015ARXIVa} is a
model compression and acceleration approach that can effectively improve the
performance of student models with the guidance of teacher models as
regularizers. The concept of knowledge distillation is well-established and has
been applied to many different tasks~\cite{Dai2021CVPR, Yang2023ICCVW,
Gu2024ICLR, Wang2022ECCVA}. As for distilling neural
fields,~\cite{Wang2024ARXIV, tan2023CVPR} proposed
to distill knowledge from a per-scene optimized NeRF-based model to a
feed-forward model, which can generalize to unseen data. Similar to knowledge
distillation,~\cite{Chen2023ARXIVa, Yu2024ARXIV,
Niemeyer2024ARXIV} proposed to leverage both the robustness of implicit
neural field and efficiency of 3DGS by training two representations jointly.
However, these prior works mainly focus on static scenes. To the best of our
knowledge, we are the first method to distill the knowledge between the
different 3D representations of human avatars with physically based rendering to
achieve real-time rendering and relighting under novel poses and novel lighting.

\subsection{Relightable Avatar}
Typical approaches for human avatar relighting often reconstruct the intrinsic
properties of humans via a multi-view capture system with controlled lighting
\cite{Debevec2000SIGGRAPH,Theobalt2007TVCG,Guo2019SIGGRAPH,Bi2021SIGGRAPH,Zhang2021SIGGRAPHa,Iwase2023CVPR,Chen2024CVPR,Saito2024CVPR,WangARXIV2025}.
In the absence of multi-view data and known illumination, R4D~\cite{Chen2022ECCVa}
jointly recovers the geometry, material properties, and lighting using a
NeRF-based representation~\cite{Zhang2021SIGGRAPHASIA,Peng2021CVPR}. However,
R4D conditions the NeRF representation on observation space encoding, making it
hard to generalize to novel poses. RANA~\cite{Iqbal2023ICCV} train a mesh
representation for multiple subjects with ground truth physical properties. Sun
et al.~\cite{Sun2023ICCV} computes the shading color via spherical Gaussian
approximations. However, those two methods do not model the visibility,
leading to the lack of shadowing effects. 

RA-X~\cite{Xu2024CVPR} leverages sphere tracing for rendering
and uses Distance Field Soft Shadow to calculate soft visibility. RA-L
\cite{Lin2024AAAI} proposed an invertible deformation field for more accurate
geometry reconstruction and a part-wise visibility MLP to model the shadowing
effects.  IA~\cite{Wang2024CVPR} leverages volumetric scattering and implements a
fast secondary ray tracing to model the visibility. The abovementioned works
achieve a high-quality estimation of intrinsic properties of human avatars.
However, they still use neural fields to represent human avatars and
employ explicit Monte Carlo integration for PBR.  Eventually, this results
in slow rendering speed, \eg\ several seconds per image.

Most recently, Li et al. \cite{Li2024ARXIV} employ 3DGS-based
ray-tracing \cite{Gao2023ARXIV} to simulate Monte Carlo integration for avatars. MeshAvatar
\cite{Chen2024ECCV} uses a mesh representation
\cite{Shen2021ARXIVa} that is amenable to efficient Monte Carlo ray-tracing.
Those two methods require dense viewpoints for training.
\cite{Zhan2024ARXIV} adopts both mesh and 3DGS as human representation and
rasterizes the mesh from all possible lighting directions to model the
visibility. GS-IA~\cite{Zhao2024ARXIV} employs 2DGS for human representation and
calculates the ambient occlusion by sampling hundreds of rays during rendering.
However, none of the concurrent work achieves real-time rendering. To be
specific, \cite{Li2024ARXIV} takes several seconds to render one image, while 
\cite{Chen2024ECCV, Zhan2024ARXIV, Zhao2024ARXIV} achieve an interactive frame
rate (5-10 FPS) using either cached occlusion probes or rasterization-based
shadow map computation.
In contrast, we achieve real-time (67 FPS) avatar relighting,
thanks to our efficient ambient occlusion probes which require only a single query
to compute shadows.

\section{Method}
\label{sec:method}
\begin{figure*}[t]
    \centering
    \includegraphics[width=\textwidth]{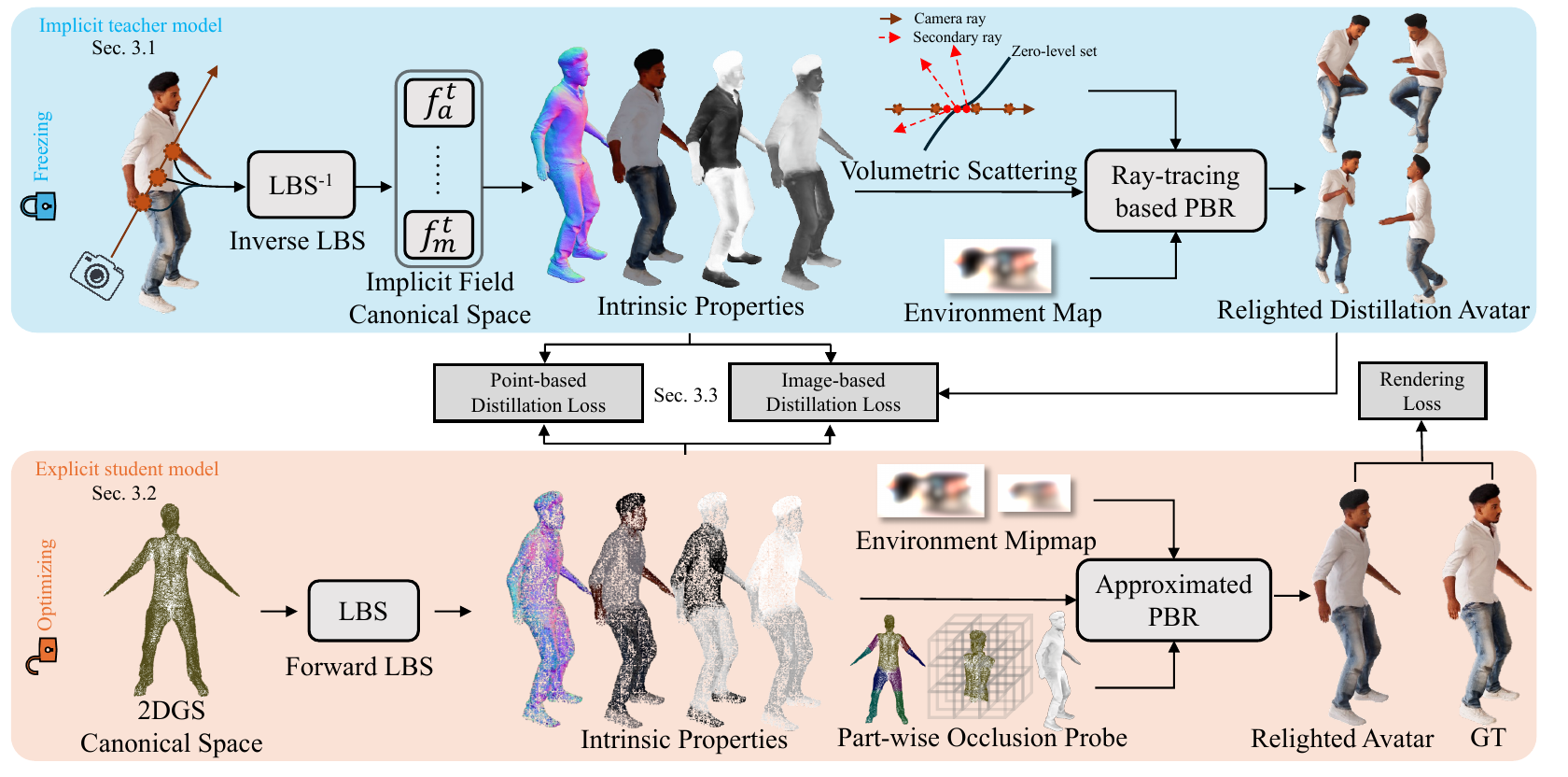}
    \vspace{-0.8cm}
    \caption{\textbf{Method overview.} Given a monocular video, we first train an implicit teacher model (\secref{sec:teacher}) via ray-tracing-based PBR to decompose the intrinsic properties, including geometry, albedo, roughness, and metallic. Then, a point-based (2DGS~\cite{Huang2024SIGGRAPH}) explicit student model (\secref{sec:student}) is optimized under the guidance of the teacher model. In order to avoid the time-consuming ray-tracing-based PBR, we adopt an approximated PBR with part-wise occlusion probes to compute the shading color and model the shadowing effects. We regularize the student model by distilling (\secref{sec:distill}) the implicit property fields from our teacher model.}
    \label{fig:method}
\vspace{-0.4cm}
\end{figure*}
In this section, we first introduce our teacher model, which is based on IntrinsicAvatar~\cite{Wang2024CVPR}. It employs Monte Carlo ray tracing and thus does not achieve real-time performance.  To enable real-time relighting, we propose a student model, which is represented as an articulated 2DGS~\cite{Huang2024SIGGRAPH} model.  We extend the appearance model of 2DGS with an approximated PBR model and propose a novel part-wise ambient occlusion model which enables real-time rendering.  The framework is described in Fig.~\ref{fig:method}

\subsection{Teacher Model}
\label{sec:teacher}
\boldparagraph{Implicit Avatar Representation} Our teacher model represents geometry and appearance of humans in canonical neural fields.  The geometry of the avatar is modeled as a neural signed-distance field (SDF). We use VolSDF~\cite{Yariv2021NEURIPS} to convert SDF into volume density for volumetric rendering.  Other appearance properties, such as roughness, metallic, and albedo, are represented using a separate neural field. More specifically, the implicit representation can be formulated as follows:
\begin{align}
\{SDF^t, r^t, m^t, \vect{a}^t \} & = f_{\{sdf, r, m, a\}}^t(\mathbf{x}_c),
\end{align}
where the superscript $t$ denotes teacher, and $f^t(\cdot)$ consists of a geometry field (for $SDF^t$) and an appearance field (for $r^t, m^t, \vect{a}^t$), represented as two individual iNGPs~\cite{Mueller2022SIGGRAPH}.  Each field takes the query point $\mathbf{x}_c$ in canonical space as input and predicts the signed distance $SDF^t$, roughness $r^t$, metallic $m^t$, or albedo $\vect{a}^t$ at that point.

In order to transform points from the observation space into the canonical space, the teacher model follows a standard skeletal-deformation-based on SMPL~\cite{Loper2015SIGGRAPH} via inverse linear-blend skinning ($\text{LBS}^{-1}$):
\begin{align}
\label{eq:lbs}
    \mathbf{x}_o = \text{LBS}(\mathbf{x}_c, \vect{\theta}) &= \sum_{i = 1}^{n_b} w_i(\mathbf{x}_c) \ \mathbf{B}_i(\vect{\theta})  
 \mathbf{x_c}, \\
 \mathbf{x}_c &= \text{LBS}^{-1}(\mathbf{x}_o, \vect{\theta}),
\end{align}
where $\{ \mathbf{B}_i \}$ are the bone transformations derived from the pose parameter $\vect{\theta}$, $w$ are the skinning weights. $\mathbf{x}_o$, $\mathbf{x}_c$ are points in the observation and canonical space, respectively.  We use Fast-SNARF~\cite{Chen2023PAMI} for the inverse LBS. %

\boldparagraph{Ray-tracing-based PBR} Our teacher model tackles the physically based rendering process as a volume scattering problem.
In this case, the ray visibility is modeled as transmittance $T(\cdot, \cdot)$. Formally, the standard equation for accumulated in-scattered radiance is as follows:
\begin{align}
\mathbf{I}^{t}(\mathbf{r})&=\int_{t_n}^{t_f} T\left(\mathbf{r}(t_n), \mathbf{r}(t)\right) \sigma_s(\mathbf{r}(t)) L_s(\mathbf{r}(t),\boldsymbol{\omega}_o) \mathrm{d} t \nonumber \\
L_s(\mathbf{x},\boldsymbol{\omega}_o)&=\int_{S^2} T\left(\mathbf{x}, \mathbf{x}+t_f^{\prime}\boldsymbol{\omega}_i\right) f_p(\mathbf{x},\boldsymbol{\omega}_o, \boldsymbol{\omega}_i) L_e^t(\boldsymbol{\omega}_i) \mathrm{d} \boldsymbol{\omega}_i \nonumber \\
T(\mathbf{x}, \mathbf{y})&=\text{exp}\left (-\int_0^{\|\mathbf{y}-\mathbf{x}\|} \sigma_t(\mathbf{x} + \frac{\mathbf{y}-\mathbf{x}}{\|\mathbf{y}-\mathbf{x}\|}t) d t\right ),
\end{align}
where $\vect{r}(t) = \vect{o}-\vect{\omega}_ot$ denotes the camera ray. $\vect{o}, \boldsymbol{\omega}_o, \boldsymbol{\omega}_i$ represent the camera center, outgoing light direction (surface to camera), and incoming light direction (surface to light), respectively. $(t_n, t_f)$ defines the near/far point for the primary ray integral. $t_f^{\prime}$ is the far point for the secondary ray integral. $L_e^t$ denotes a learnable environment map. $f_p, \sigma_s, \sigma_t$ are phase function, scattering coefficient, and attenuation coefficient, which are parameterized by the implicit intrinsic properties $r^t, m^t, \vect{a}^t$. The overall volume scattering process requires sampling and evaluating hundreds of points and secondary rays to render one pixel $\mathbf{I}^{t}(\mathbf{r})$, which is quite time-consuming.  In this work, we distill the knowledge from this implicit, ray-tracing-based teacher model to an explicit student model with an approximated PBR to achieve real-time rendering at inference time, while preserving the high-fidelity rendering results.
\subsection{Student Model}
\label{sec:student}
\boldparagraph{Explicit Avatar Representation}
We extend the 2DGS~\cite{Huang2024SIGGRAPH} representation by adding the intrinsic attribute to each Gaussian primitives $\mathcal{P}_i$:
\begin{align}
\{\mathcal{P}_i\} = \left\{\left (\vect{\mu}_{c,i}^s, \mathbf{q}_{c,i}^s, \mathbf{s}^s_i, o^s_i, \mathbf{c}^s_i, \mathbf{a}^s_i, r^s_i, m^s_i \right ) 
| i \in \left [0, N \right ) \right\},
\end{align}
where the superscript $s$ represents student, and $\vect{\mu}_{c}^s, \mathbf{q}_{c}^s, \mathbf{s}^s, o^s, \mathbf{c}^s, \mathbf{a}^s, r^s, m^s$ denote mean, quaternion rotation, scale, opacity, radiance, albedo, roughness, and metallic, respectively. The normal of a 2DGS is defined by the last column of its rotation matrix, \ie $\vect{n^s_c}=\mathbf{R}(\mathbf{q}_c^s)_{:,3}$.
Similar to our teacher model, we define our 2DGS in canonical space.  We apply forward LBS to obtain Gaussians in observation space:
\begin{align}
\vect{\mu}_o^s &= \text{LBS}(\vect{\mu}_c^s, \vect{\theta}), \\
\mathbf{R}(\mathbf{q}_o^s) &= \text{LBS}(\mathbf{R}(\mathbf{q}_c^s), \vect{\theta}),
\end{align}
where both the mean $\vect{\mu}_c^s$ and rotation $\mathbf{q}_c^s$ are transformed from the canonical space to the observation space $\vect{\mu}_o^s$, $\mathbf{q}_o^s$. After transforming Gaussian primitives to the observation space, we use split-sum approximation to compute the shading color for each Gaussian primitive.

\boldparagraph{Approximated PBR} The exact PBR color of a Gaussian is the integral of the multiplication of the BRDF $f_r$, incident radiance $L_i$, visibility $V$, and foreshortening term $\mathbf{n} \cdot \vect{\omega}_{i}$ over the hemisphere defined by the normal $\vect{n}$ of the Gaussian:
\begin{align}
\label{eq:gs_pbr}
\mathbf{c}^{gs}(\mathcal{P}, \vect{\omega}_o) &= \int_{\Omega} f_{r}\left(\vect{\omega}_{i}, \vect{\omega}_{o}\right) L_{i}\left( \vect{\omega}_{i}\right) V\left(\vect{\mu},\vect{\omega}_{i}\right) \mathbf{n} \cdot \vect{\omega}_{i} 
\mathrm{d} \vect{\omega}_{i} \nonumber \\
&\approx \text{AO}(\vect{\mu},\vect{n}) \int_{\Omega} f_{r}\left(\vect{\omega}_{i}, \vect{\omega}_{o}\right) L_{i}\left( \vect{\omega}_{i}\right) \mathbf{n} \cdot \vect{\omega}_{i} \mathrm{d} \vect{\omega}_{i} \nonumber \\
&= \text{AO}(\vect{\mu},\vect{n})\left [ \underbrace{L_d\left ( \mathbf{n} \right)}_{\text{Diffuse}} + \underbrace{L_s \left ( \mathbf{n}, \vect{\omega}_{i}, \vect{\omega}_{o} \right)}_{\text{Specular}}\right]
\end{align}
where we adopt Cook-Torrance BRDF so that the equation is separated into a diffuse $L_d$ and a specular $L_s$ part. Note that we approximate the visibility by leveraging an Ambient Occlusion (AO) term:
\begin{align}
\label{eq:ao}
\text{AO}(\vect{\mu},\vect{n}) = \frac{1}{\pi}\int_{\Omega}V\left(\vect{\mu},\vect{\omega}_{i}\right) \mathbf{n} \cdot \vect{\omega}_{i} \mathrm{d} \vect{\omega}_{i}.
\end{align}
We precompute the part-wise ambient occlusion and store it in the occlusion probe grid to avoid time-consuming ray tracing during rendering. The details are introduced in the last part of this section. 

The diffuse part $L_d$ and the specular part $L_s$ are formulated below:
\begin{align}
\label{eq:pre_diffuse}
&L_d\left ( \mathbf{n} \right) = \frac{(1-m)\mathbf{a}}{\pi} \int_{\Omega} L_{i}\left( \vect{\omega}_{i}\right)\mathbf{n} \cdot \vect{\omega}_{i} \mathrm{d} \vect{\omega}_{i} \\
& L_s \left ( \mathbf{n}, \vect{\omega}_{i}, \vect{\omega}_{o} \right) = \int_{\Omega} f_s\left ( \mathbf{n}, \vect{\omega}_{i}, \vect{\omega}_{o} \right) L_{i}\left( \vect{\omega}_{i}\right)\mathbf{n} \cdot \vect{\omega}_{i} \mathrm{d} \vect{\omega}_{i} \nonumber \\
\label{eq:pre_specular}
& \approx \underbrace{\int_{\Omega}L_{i}\left( \vect{\omega}_{i}\right) \mathbf{n} \cdot \vect{\omega}_{i} \mathrm{d} \vect{\omega}_{i}}_{\text{Pre-filtered environment mipmap}} \cdot \underbrace{\int_{\Omega}f_s\left ( \mathbf{n}, \vect{\omega}_{i}, \vect{\omega}_{o} \right) \mathbf{n} \cdot \vect{\omega}_{i} \mathrm{d} \vect{\omega}_{i} }_{\text{Pre-computed BRDF}}
\end{align}

We adopt the split-sum approximation~\cite{Karis2013SIGGRAPH} for the specular part, resulting in two separate integrals. All three integrals from diffuse and specular parts can be precomputed and stored in look-up tables. Also, the specular BRDF takes the intrinsic properties $\{\vect{n}, \vect{a}, r, m\}$ from each Gaussian into account:
\begin{align}
f_s &= \frac{D(\boldsymbol{n},\boldsymbol{h} ; r) F\left(\boldsymbol{\omega}_o, \boldsymbol{h} ; \boldsymbol{a}, m\right) G\left( \boldsymbol{n}, \boldsymbol{\omega}_i, \boldsymbol{\omega}_o; r\right)}{\left(\boldsymbol{n} \cdot \boldsymbol{\omega}_i\right) \left(\boldsymbol{n} \cdot \boldsymbol{\omega}_o\right)},
\end{align}
where $D$, $F$, and $G$ denote microfacet normal distribution function, Fresnel term, and geometry term, respectively. 

Finally, we follow the rasterization process of 2DGS to render the image $\vect{I}^{s}$ based on the per-Gaussian PBR color $\vect{c}^{gs}$ we calculated from Eq.~\eqref{eq:gs_pbr}:
\begin{align}
\label{eq:rast}
\mathbf{I}^{s}(\mathbf{r},\mathbf{c}^{gs})=\sum_{i=1} \mathbf{c}^{gs}_i o_i \hat{\mathcal{G}}_i(\mathbf{u}(\mathbf{r})) \prod_{j=1}^{i-1}\left(1-o_j \hat{\mathcal{G}}_j(\mathbf{u}(\mathbf{r}))\right),
\end{align}
where $\vect{r}$ is the camera ray, $\vect{u}(\cdot)$ returns the $uv$ coordinate of the intersection point between the camera ray and the Gaussian primitives. $\hat{\mathcal{G}}(\cdot)$ is a bounded 2D Gaussian density function.\\
\boldparagraph{Part-wise Occlusion Probe} Inspired by GS-IR~\cite{Liang2024CVPR}, we leverage spherical harmonics (SH) coefficients to store occlusion information.  Different from \cite{Liang2024CVPR}, where binary occlusion cubemaps are directly converted into SH, we convert the \textit{pre-convolved} ambient occlusion to SH.  Pre-convolved ambient occlusion is much smoother compared to binary occlusion maps, and thus can be better captured by SH which represents low-frequency signals better.  It also allows us to compute shadows for a pixel with a single query, which is crucial for real-time rendering.  Formally, we first generate a 3D grid in canonical space for each body part.  Then, we compute and store SH coefficients on each grid point.  For each body part $p$, the SH coefficient $f_{l m}^{p}(\vect{\mu})$ at point $\vect{\mu}$ is calculated as follows: 
\begin{align}
f^{p}_{l m}(\vect{\mu})=\int_{S^2} \text{AO}^p\left (\vect{\mu},\vect{\omega} \right ) Y_{l}^{m}(\boldsymbol{\omega}) d \boldsymbol{\omega},
\end{align}
where $\{Y_{l}^{m}\}$ denotes the basis of SH, and $\text{AO}^p\left (\vect{\mu},\vect{\omega} \right )$ is the ambient occlusion for body part $p$, which is obtained by rasterizing six times at point $\vect{\mu}$ to form a occlusion cubemap, then convolve it with the clamped cosine lobe via Eq.~\eqref{eq:ao}. After converting the ambient occlusion to SH, it can be recovered during rendering:
\begin{align}
\hat{\text{AO}}^{p}(\vect{\mu}, \vect{\omega})=\sum_{l=0}^{d e g} \sum_{m=-l}^l f_{l m}^{p}(\vect{\mu}) Y_{l}^{m}(\vect{\omega}),
\end{align}
where $deg$ is the degree of SH. To model the shadowing effect caused by body articulation, we transform the point $\vect{\mu}_o$ and the normal $\vect{n}_o$ from the Gaussian primitive in observation space to the canonical space of each body part. Then, we query the $\hat{\text{AO}}^{p}$ for each part and multiply them together to get the final ambient occlusion:
\begin{align}
\label{eq:hat_ao}
\hat{\text{AO}}(\vect{\mu}_o, \vect{n}_o) = \prod_{p=1}^{N_p} \hat{\text{AO}}^{p}(\mathbf{B}_p(\vect{\theta})^{-1}\vect{\mu}_o, \mathbf{B}_p(\vect{\theta})_{1:3,1:3}^{-1}\vect{n}_o),
\end{align}
where $\mathbf{B}_p(\vect{\theta})$ denotes the bone transformation of part $p$ given body pose $\vect{\theta}$.

\subsection{Objectives}
\label{sec:distill}
\boldparagraph{Point-based Distillation Loss} We distill the knowledge from our teacher model to each Gaussian primitive by querying the corresponding neural fields for different intrinsic properties:
\begin{align}
L_{distill}^{p} =\frac{1}{N} \sum_{i=1}^{N} l_{*} \left (f_{adapt}(*^s_i),  f^t_* \left (\boldsymbol{\mu}^s_{c,i} \right ) \right ),
\end{align}
where $*$ denotes $\{r,m,\vect{a},\vect{n}_c \}$. Notice that we adopt L1 loss for $l_r,l_m,l_a$ and cosine similarity loss for $l_n$. Also, we calculate the gradient of SDF as the normal field, \ie $f_n = \nabla f_{sdf} $. We introduce an adapt layer $f_{adapt}$, which contains a learnable scale and bias to cope with the misalignment between ray-tracing PBR (teacher) and split-sum PBR (student). Besides, we regularize the 2D primitives to align with the zero-level set of the teacher's SDF:
\begin{align}
L_{distill}^{sdf} =\frac{1}{N} \sum_{i=1}^{N}  \left \|  f^t_{sdf} \left (\boldsymbol{\mu}^s_{c,i} \right ) \right \|_2,
\end{align}
\boldparagraph{Image-based Distillation Loss}
We also introduce an image-based distillation loss to regularize the predicted intrinsic properties of the student model in image space. The teacher model renders properties by replacing the radiance with the corresponding properties in the volume rendering equation:
\begin{align}
\vect{I}^{t}(\mathbf{r},*)&=\int_{t_n}^{t_f} T\left(\mathbf{r}(t_n), \mathbf{r}(t)\right) \sigma_t(\mathbf{r}(t)) f^t_*(\text{LBS}^{-1}(\mathbf{r}(t))) \mathrm{d} t
\end{align}
Similarly, we replace the $\mathbf{c}^{gs}$ in Eq.~\eqref{eq:rast} to render intrinsic properties for the student model. The image-based distillation loss is calculated as follows:
\begin{align}
L_{distill}^{i} =\frac{1}{|\mathcal{R}|} \sum_{\boldsymbol{r} \in \mathcal{R}} l_* \left (\mathbf{I}^{t}(\mathbf{r},*), f_{adapt}( \mathbf{I}^{s}(\mathbf{r},*)) \right ),
\end{align}
where $\mathcal{R}$ denotes the set of the camera ray of the image.

\boldparagraph{Rendering Loss} We supervise our student model with ground truth images:
\begin{align}
L_{r} =   L_{rgb}\left (\mathbf{I}^{s}(\mathcal{R},\mathbf{c}^{gs}),\mathbf{I}^{gt}_{rgb} \right ) + L_{mask}\left (\mathbf{I}^{s}(\mathcal{R},1),\mathbf{I}^{gt}_{mask} \right ),
\end{align}
where $\mathbf{I}^{gt}_{rgb}$ and $\mathbf{I}^{gt}_{mask}$ denotes ground truth images and masks separately. $L_{rgb}$ consists of L1 and LPIPS loss, and $L_{mask}$ is L1 loss.

\boldparagraph{Distillation Avatar} In addition to the ground truth training image, we also sample some poses from AIST\cite{li2021learn} and RANA\cite{Iqbal2023ICCV} datasets to allow the teacher model to render additional pseudo ground truth images as distillation avatar for the student to learn. This is crucial for the implicit teacher model to distill the inductive bias, \eg the density of Gaussian primitives around joints, the interpolation ability of MLP, to the explicit student model to help the student generalize well to out-of-distribution poses during animation. 

\boldparagraph{Regularization Loss} We regularize the intrinsic properties $\{r,m,\vect{a}\}$ via a bilateral smoothness term~\cite{Gao2023ARXIV}. Besides, we also incorporate the depth distortion and normal consistency loss from 2DGS, an anisotropy regularizer from PhysGaussian \cite{Xie2023ARXIV}, as well as a normal orientation loss. The final loss is a linear combination of the losses with the corresponding weights. See \suppmat for more details.

\section{Experiments}
\label{sec:experiments}

\subsection{Datasets and Metrics}

\boldparagraph{RANA Dataset \cite{Iqbal2023ICCV}} We use this synthetic dataset to quantitatively and qualitatively assess the reconstructed avatar under novel poses and novel illumination conditions.  Following the same setting of IntrinsicAvatar \cite{Wang2024CVPR}, we select 8 subjects from the RANA dataset.
The dataset provides ground truth albedo, normal, and relighted images for evaluation.  We adhere to Protocol A, where the training set contains subjects in an A-pose, rotating in front of the camera under unknown illumination. The test set consists of images of the same subjects in random poses under novel illumination.

\boldparagraph{PeopleSnapshot Dataset \cite{Alldieck2018CVPR}} This dataset is a real-world dataset, which consists of subjects consistently holding A-pose while rotating in front of the camera under natural illumination.  Following~\cite{Wang2024CVPR}, we use refined pose estimations from \cite{Jiang2023CVPR}.  This dataset is used only for qualitative evaluation.  

\boldparagraph{Metrics} Relighting quality and albedo are measured via PSNR, SSIM, and LPIPS. We assess geometry using Normal Error (degree). We report the frame rate per second (FPS) for rendering speed. See \suppmat for more details.

\subsection{Baselines}
We choose two state-of-the-art methods as baselines for comparison, \ie\ R4D~\cite{Chen2022ECCVa} and IntrinsicAvatar (IA)~\cite{Wang2024CVPR}. Notice that IntrinsicAvatar~\cite{Wang2024CVPR} is the most recent physically based inverse rendering method for human avatars with publicly available training code. RelightableAvatar \cite{Xu2024CVPR} and RANA \cite{Iqbal2023ICCV} do not fully release their training guidance at present. Thus, we do not take them as our baselines.

\begin{figure*} 
    \centering
    \includegraphics[width=\textwidth,trim=0 7 0 0,clip]{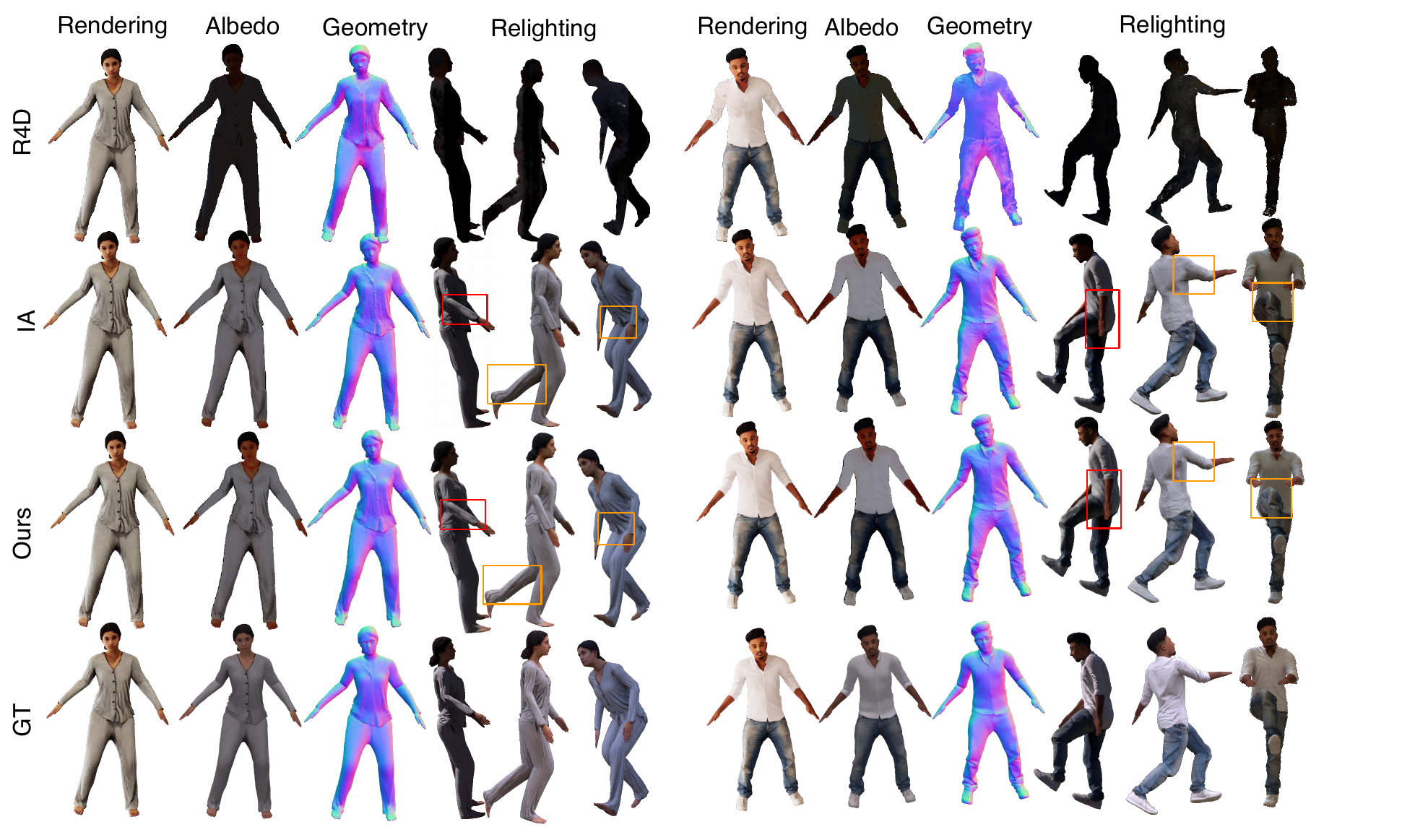}
    \vspace{-0.9cm}
    \caption{\textbf{Qualitative Relighting and Decomposition Comparison on RANA.} \textbf{\textcolor[RGB]{255,0,0}{Red}} bounding boxes: the noisy shading results caused by the artifacts of geometry. \textbf{\textcolor[RGB]{255,125,0}{Orange}} bounding boxes: incorrect shading results caused by the wrongly estimated materials/visibility. 
    }
    \label{fig:main_results}
\vspace{-0.5cm}
\end{figure*}

\subsection{Relighting Comparisons}
\tableRelighting
In the pipeline introduced in \secref{sec:method}, we have used forward shading
(denoted as Ours-F), where we first compute shading color via \equref{eq:gs_pbr}
and then rasterize it via \equref{eq:rast}.  In this section,  we also explore
an alternative using deferred shading (denoted as Ours-D), where we first
rasterize the albedo, occlusion, roughness, and metallic map in screen space,
and then perform the shading in screen space.

As shown in \tabref{tab:relighting}, we compare our method with our teacher
model and R4D.  We achieve around 300x faster rendering speed compared to NeRF-based methods.  In the meantime, we obtain comparable relighting results
or even surpass our teacher model on SSIM and PSNR. We also note that the
deferred shading version of our method (Ours-D) achieves better LPIPS,
at the cost of significantly slower rendering speed, dropping from 67 FPS to 33
FPS.  This is because the overhead of deferred shading is dominated
by the number of pixels in the image, while the overhead of forward shading is mainly
determined by the number of Gaussian primitives, which is much fewer than the
number of pixels.  We use forward shading as our default method in the following
experiments.

Qualitative results are shown in \figref{fig:main_results}. R4D fails to produce
reasonable results due to its inability to generalize to novel poses.  IA tends
to produce high-frequency noise in certain areas
(\textcolor[RGB]{255,0,0}{Red} bounding boxes) due to the utilization
of iNGP. Moreover, the volumetric scattering-based teacher model may sample secondary rays inside the surface compared to a surface-based student model, leading to a darker shadowing effect. Also, the limited sample counts of IA may result in noisy or wrongly estimated materials. The \textcolor[RGB]{255,125,0}{orange} bounding boxes in \figref{fig:main_results} confirm that.

\begin{figure*} 
    \centering
    \includegraphics[width=\textwidth,trim=0 7 0 0,clip]{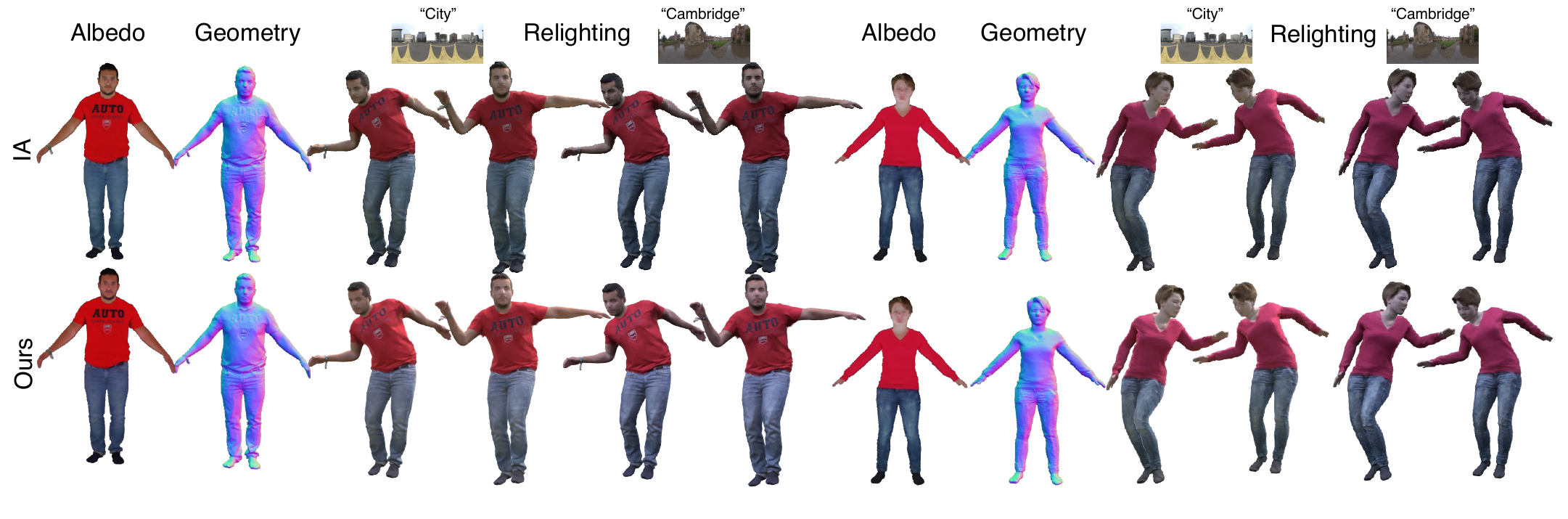}
    \vspace{-0.9cm}
    \caption{\textbf{Qualitative Relighting and Decomposition comparison on PeopleSnapshot.} IA produces noisy face relighting due to geometric artifacts and noisy Monte Carlo sampling, whereas our model delivers high-quality relighting results with sharp boundaries.
    }
    \label{fig:ps_results}
\vspace{-0.5cm}
\end{figure*}

In addition, we show the results on the real-world dataset in \figref{fig:ps_results}. Similarly, IA suffers from noises caused by iNGP and Monte Carlo estimation, leading to blurry and noisy facial relighting results. On the contrary, our model produces smoother geometry, thanks to 2DGS, while the split-sum-based appearance model does not suffer from noises that are common in Monte Carlo estimation.
\vspace{-0.2em}
\subsection{Intrinsic Properties Comparisons}
\vspace{-0.2em}
\tableProperty
We also compare with R4D and IntrinsicAvatar for
the task of intrinsic property decomposition on the RANA dataset. As shown in
the \tabref{tab:decomposition}, our method outperforms our teacher model in terms
of normal consistency. This disparity becomes more visible in qualitative
comparisons shown in \figref{fig:main_results}: IA tends to have
texture details baked into geometry, whereas our model successfully
keeps meaningful wrinkles while discarding high-frequency noise during
distillation, thanks to the smoothness prior from 2DGS.
For albedo estimation, our student model achieves comparable accuracy to that of the teacher model.
Although increasing the albedo distillation loss enables the student model to closely match the teacher's albedo, this exact replication proves to be suboptimal for relighting under our approximated PBR pipeline for student. Instead, our objective prioritizes improved relighting performance over precise intrinsic decomposition.
\vspace{-0.3em}
\subsection{Ablation Study}
\vspace{-0.2em}
\begin{figure} 
    \centering
    \vspace{-0.3cm}
    \includegraphics[width=\linewidth]{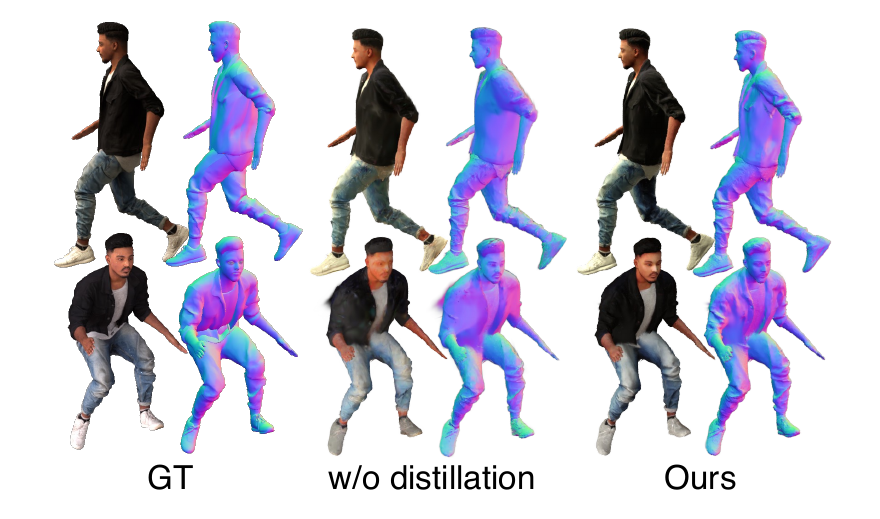}
    \vspace{-0.9cm}
    \caption{\textbf{Ablation study for distillation.} Our model produces fine-grained geometry with the help of our proposed knowledge distillation strategy, leading to high-quality relighting results. 
    }
    \label{fig:ablation_distillation}
\vspace{-0.5cm}
\end{figure}

\begin{figure} 
    \centering
    \includegraphics[width=\linewidth]{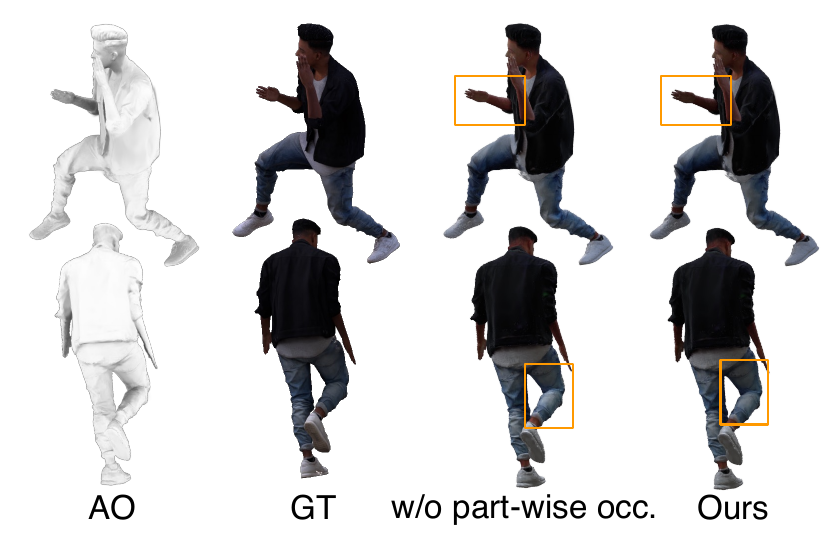}
    \vspace{-0.9cm}
    \caption{\textbf{Ablation study for part-wise occlusion probe.} Our part-wise occlusion probe successfully models the shadow effects between different body parts, resulting in similar relighting results with ground truth images.
    }
    \vspace{-0.8cm}
    \label{fig:ablation_occ}

\end{figure}

We ablate several of our design choices. We use subject 01 from the RANA dataset for this ablation study. 

\tableAblation

\boldparagraph{Knowledge distillation} As depicted in \tabref{tab:ablation},
knowledge distillation serves as an efficient regularization term that
drastically improves the relighting quality. This is also confirmed by
qualitative results in \figref{fig:ablation_distillation}, where only optimizing
the explicit representation itself can not produce satisfying geometry and
easily gets stuck into local optima, leading to noisy relighting results. Moreover, as shown in the second row of \tabref{tab:ablation}, the additional distillation avatar rendered from sampled poses successfully distills the inductive bias from the implicit teacher model to the explicit one, making student model generalize well to out-of-distribution novel poses. This is also confirmed by \figref{fig:supp_ood} in \suppmat.

\boldparagraph{Part-wise occlusion probes} The relighting quality decreases
if we turn off our part-wise occlusion probes, as quantitatively indicated by the second
row of \tabref{tab:ablation}. The \textcolor[RGB]{255,125,0}{orange} bounding
boxes in \figref{fig:ablation_occ} serve as visual evidence for the necessity of the proposed part-wise occlusion probes.
The part-wise occlusion probes capture shadows on the forearm and
between the upper and lower parts of the leg, resulting in relighting results that are more consistent with ground truth images.

\section{Conclusion}
\label{sec:conclusion}
In this paper, we present \methodname, which reconstructs relightable human avatars that support real-time rendering from monocular videos.
We represent humans as 2DGS and adopt an approximated PBR to compute shading color.  We show that novel part-wise ambient occlusion probes are curical to achieving realistic shadows with real-time performance.  We also demonstrated that it is necessary to distill and regularize our model with a ray-tracing-based teacher model to achieve high-quality results.  Overall, our model achieves comparable results with our teacher model while being hundreds of times faster at inference, achieving a 67 FPS rendering speed under unseen environment lighting and unseen poses.

\boldparagraph{Acknowledgement} Zeren Jiang is supported by Clarendon Scholarship. Shaofei Wang and Siyu Tang acknowledge SNSF grant 200021 204840.

{
    \small
    \bibliographystyle{ieeenat_fullname}
    \bibliography{main, bibliography,bibliorgraphy_custom}
}

\clearpage
\maketitlesupplementary

In this \textbf{supplementary document}, we provide additional materials to supplement our main submission. 
The \textbf{code} is available here for research purposes: \rurl{github.com/jzr99/DNF-Avatar}

\section{Implementation Details}

\subsection{Final Objectives}
In addition to the losses introduced in our manuscript, we also adapt the following loss during distillation. The final loss is a linear combination of the losses with the corresponding weights.

\noindent\textbf{Material Smoothness Loss.} We regularize the intrinsic properties $\{r,m,\vect{a}\}$ via a bilateral smoothness term\cite{Gao2023ARXIV}, which prevents the material properties from changing drastically in areas with smooth colors:
\begin{align}
\mathcal{L}_{\text {smooth}}=\|\nabla \mathbf{I}^s\left(\mathcal{R},*\right)\| \exp \left(-\left\|\nabla \mathbf{I}^{gt}_{rgb}\right\|\right),
\end{align}
where $\mathbf{I}^s\left(\mathcal{R},*\right)$ are rasterized material maps. $*$ denotes $\{r,m,\vect{a}\}$. $\mathbf{I}^{gt}_{rgb}$ represents ground truth images.

\noindent\textbf{Anisotropy Regularization Loss.} We adopt the loss from \cite{Xie2023ARXIV} for 2DGS:
\begin{align}
\mathcal{L}_{\text {aniso }} & = \frac{1}{N} \sum_{i=1}^{N} \max \left\{\max \left(\mathbf{s}_i^s\right) / \min \left(\mathbf{s}_i^s\right), r\right\}-r,
\end{align}
where $\mathbf{s}^s_i$ is the scaling of 2DGS. This loss constrains the ratio between the length of two axes of 2DGS that to not exceed predefined value $r$. We set $r=3$ to prevent the Gaussian primitives from becoming threadlike, which alleviates the geometric artifacts under novel poses.

\noindent\textbf{Normal Orientation Loss.} 
Ideally, normals of visible 2D Gaussian primitives should always face toward the camera. To enforce this, we employ the normal orientation loss~\cite{Verbin2022CVPR}:
\begin{align}
L_{orient} =\frac{1}{|\mathcal{R}|} \sum_{\boldsymbol{r} \in \mathcal{R}} \left \| \max \left (-\vect{\omega}_{o,r}\cdot \mathbf{I}^{s}(\mathbf{r},\mathbf{n}^s_o), 0 \right) \right \|_1 ,
\end{align}
where $\vect{\omega}_{o,r}$ denotes the outgoing light direction (surface to camera) for ray $\vect{r}$. $\mathbf{I}^{s}(\mathbf{r},\mathbf{n}^s_o)$ denotes the rasterized world-space normal for ray $\vect{r}$.

\noindent\textbf{Environment Map Distillation Loss.} 
In addition to the distillation loss between the two avatar representations, we also regularize the environment map of our student model with the one of our teacher model:
\begin{align}
L_{distill}^{env} =\frac{1}{|\mathcal{S}^2|} \sum_{\vect{\omega} \in \mathcal{S}^2} \left \| L_e^t(\vect{\omega}) - L_e^s(\vect{\omega}) \right \|_2 ,
\end{align}
where $L_e^t$ denotes a spherical-gaussian-based environment map from our teacher model, and $L_e^s$ represents a cubemap-based environment map from our student model. $\mathcal{S}^2$ is all possible lighting directions.

\noindent\textbf{Depth Distortion and Normal Consistency.} 
Following 2DGS\cite{Huang2024SIGGRAPH}, we apply the depth distortion loss and normal consistency loss to concentrate the weight distribution along the rays and make the 2D splats locally align with the actual surfaces:
\begin{align}
L_{dist} &=\frac{1}{|\mathcal{R}|} \sum_{\mathbf{r} \in \mathcal{R}} \sum_{i,j}^{N} w_i(\mathbf{r})w_j(\mathbf{r}) \left \| z_i(\mathbf{r}) - z_j(\mathbf{r})  \right \|_1 ,\\
L_{nc} &= \frac{1}{|\mathcal{R}|} \sum_{\mathbf{r} \in \mathcal{R}}\sum_{i}^{N}w_i(\mathbf{r})(1-\mathbf{n}_i^\intercal \mathbf{N}(\mathbf{r})),
\end{align}
where $w_i(\mathbf{r})= o_i \hat{\mathcal{G}}_i(\mathbf{u}(\mathbf{r})) \prod_{j=1}^{i-1}\left(1-o_j \hat{\mathcal{G}}_j(\mathbf{u}(\mathbf{r}))\right)$ is the blending weight for $i$th 2D splat along the ray $\mathbf{r}$, and $z_i$ is the depth of the intersection point. $\mathbf{N}$ is the normal derived from the depth map.

\subsection{Training Details}
The teacher model is trained first and then frozen during distillation.  We apply the marching cube algorithm to extract the mesh from the implicit teacher model and initialize the 2DGS with a sampled subset from the vertexes of the mesh. Similar to~\cite{Zhan2024ARXIV}, during distillation, we periodically densify and prune the 2DGS with the initial sampled vertex to regularize the density of the 2DGS. Following IA~\cite{Wang2024CVPR}, we employ a two-stage training strategy during distillation. We train a total of 30k iterations with distillation loss applied. We apply a color MLP~\cite{Qian2024CVPR} to estimate the radiance in the first 20k iterations, while we employ both color MLP and PBR rendering loss for the rest of the iterations. Note that the color MLP is only used during training, which helps regularize the geometry of the Gaussians. As for the precomputation of occlusion probes, we separate the human avatar into 9 parts based on the skinning weights, and precompute the part-wise occlusion probes after the first 20k iterations.

During rendering, we adopt the standard gamma correction to the rendered image from linear RGB space to sRGB space and then clip it to [0, 1]. To stay consistent with R4D~\cite{Chen2022ECCVa} and IA~\cite{Wang2024CVPR}, we calibrate our albedo prediction to the range [0.03, 0.8], which prevents the model from predicting zero albedo for near-black clothes.

\section{Additional Experimental Results}
\subsection{Metrics}
For synthetic datasets, we assess several metrics:

\textbf{Relighting PSNR/SSIM/LPIPS:} We evaluate standard image quality metrics for images rendered under novel poses and illumination conditions.

\textbf{FPS:} We report the rendering frame rate per second for the $540 \times 540$ resolution images on a single NVIDIA RTX 4090 GPU.

\textbf{Normal Error:} This metric measures the error (in degrees) between the predicted normal images and the ground-truth normal images.

\textbf{Albedo PSNR/SSIM/LPIPS:} We use standard image quality metrics to evaluate albedos rendered from training views. Since there is inherent ambiguity between the estimated albedo and light intensity, we align the predicted albedo with the ground truth, following \cite{Zhang2021SIGGRAPHASIA}.

For real-world datasets, \ie\ PeopleSnapshot, we provide qualitative results, showcasing novel views and pose synthesis under new lighting conditions.

\subsection{Additional Qualitative Results}
\begin{figure*} 
    \centering
    \includegraphics[width=400pt,trim=0 7 0 0,clip]{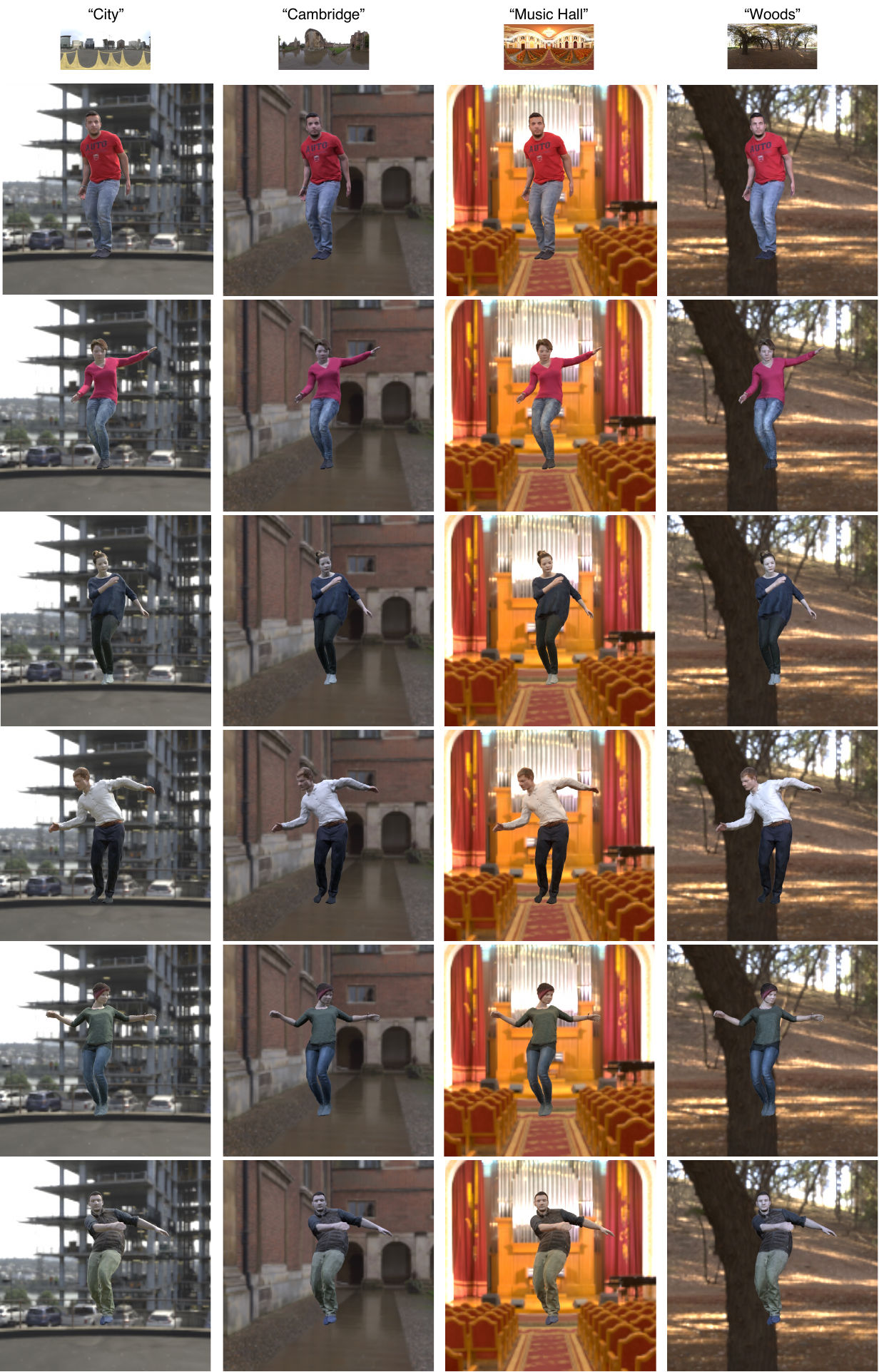}
    \caption{\textbf{Qualitative Relighting on PeopleSnapshot Dataset.} 
    }
    \label{fig:supp_ps_results}
\vspace{-0.5cm}
\end{figure*}

We show additional qualitative relighting results on the PeopleSnapshot dataset in \figref{fig:supp_ps_results}. All of the subjects are rendered under novel poses and novel illuminations.

\subsection{Additional Quantitative Results}

The per-subject and average metrics of R4D, IA, Ours-D, and Ours-F are reported in \tabref{tab:metric_all_rana}. Note that the only difference between Ours-D and Ours-F is in the inference stage, so they share the same intrinsic properties.

\subsection{Additional Ablation Study for Distillation}
\begin{figure} 
    \centering
    \includegraphics[width=200pt]{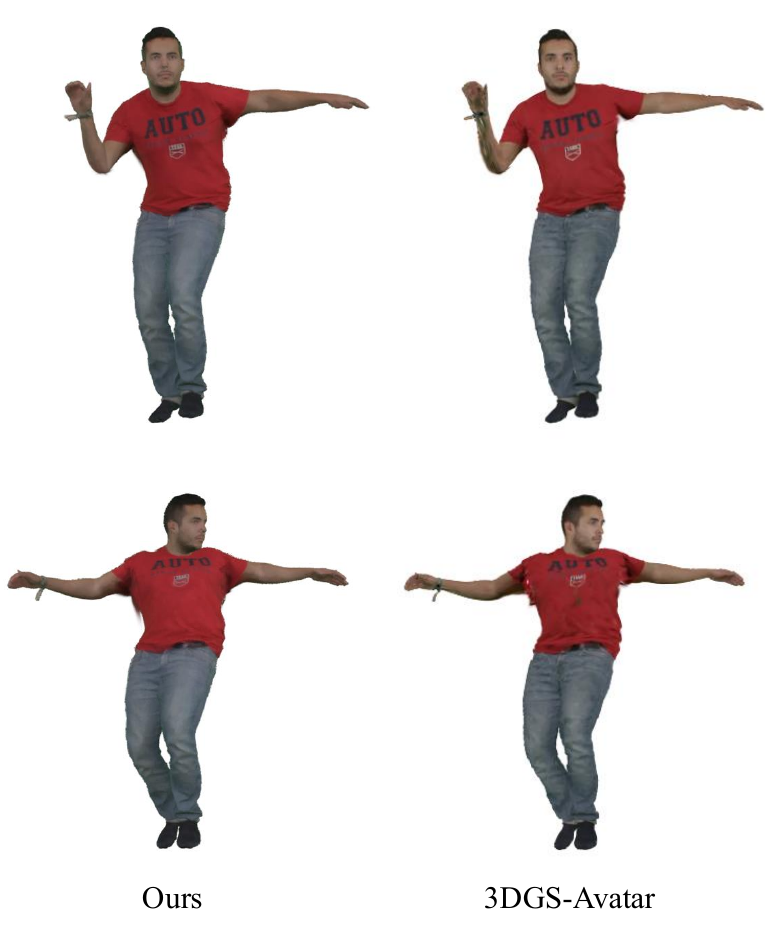}
    \caption{\textbf{Implicit bias helps pose generalization.} 
    Under limited training pose variation, the bias imposed by our implicit teacher model helps our student model to achieve reasonable rendering on out-of-distribution poses (left). In comparison, the state-of-the-art 3DGS-based avatar model~\cite{Qian2024CVPR} tends to fail on out-of-distribution poses, especially around joints (right).
    }
    \label{fig:supp_ood}
\end{figure}

As shown in \tabref{tab:supp_distill}, we ablate the proposed distillation objectives on subject 01 of the RANA dataset. dist., i-dist., and p-dist. represent distillation, image-based distillation, and point-based distillation, respectively. When distillation is disabled, 2DGS itself cannot produce satisfying geometry, leading to poor relighting results. While image-based distillation successfully distills the knowledge from the training view, point-based distillation further improves the performance by distilling knowledge in both visible and occluded areas.  We also note that the bias from the implicit teach model (smooth interpolation of density and color in regions not seen during training) helps reducing artifacts in our student model.  We compare our model with a pure explicit 3DGS-based avatar model~\cite{Qian2024CVPR} and show that such explicit representation struggles to generalize to out-of-distribution joint angles, while our model achieves reasonable results, thanks to the smoothness bias distilled from the teacher model (Fig.~\ref{fig:supp_ood}).
\tablesuppdistill

\subsection{Rendering Speed}

\tableFPS

As shown in \tabref{tab:fps}, we test the performance for each component of our
PBR pipeline. The test is done with a $540 \times 540$ resolution using around
$70000$ Gaussian primitives. The deferred shading version is bounded by the
shading time, which scales linearly with the number of pixels. In comparison,
for forward shading, the shading module itself is very fast, while querying
part-wise occlusion probes becomes the bottleneck of performance.  The
bottleneck of part-wise occlusion probes is governed by the number of Gaussian
primitives.  In addition, we assume the environment map remains unchanged for a
single animation sequence so that the precomputation time (around 10ms per
environment map) for the 
Equ.~(10) and Equ.~(11) 
is ignored.  However, our forward shading pipeline can still achieve around 40 FPS,
even if we take this precomputation into account.

\section{Additional Discussion}
\begin{figure} 
    \centering
    \includegraphics[width=0.8\linewidth]{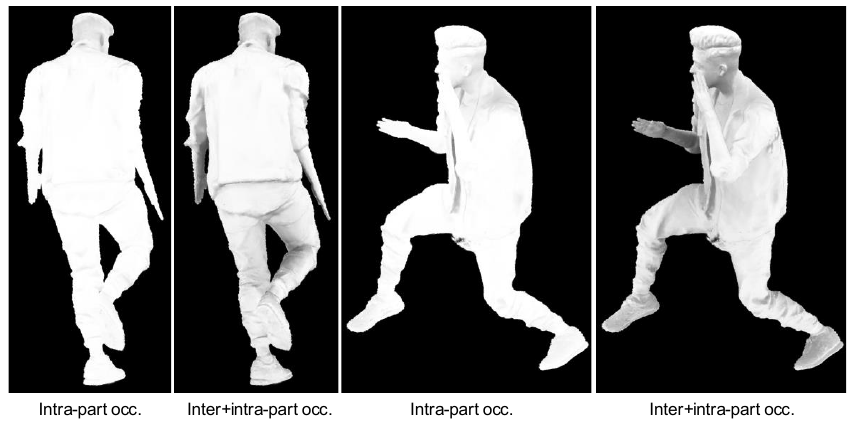}
    \caption{Visualization of intra/inter-part occlusion.}
    \vspace{-0.3cm}
    \label{fig:rebuttal_occ}
\end{figure}
\subsection{Clarification for Part-wise Occlusion} Ambient occlusion is calculated on the fly during animation.  The idea is that each body part is rigid, and thus we can pre-compute its occlusion probes in canonical space of each part.  The pre-convolution (Eq.~\eqref{eq:ao}) ensures that a single query is sufficient to obtain ambient occlusion at test time.
For a single Gaussian in observation space, we transform it to the canonical space of $N_p$ body parts according to per-part rigid transforms. By querying partwise ambient occlusion probes in canonical space, we obtain $N_p$ ambient occlusion values, and the product of these values is the final ambient occlusion (Eq.~\eqref{eq:hat_ao}). 
We also present Fig.~\ref{fig:rebuttal_occ}, where intra-part occlusion (calculated when Gaussian was transformed to the canonical space of its own part) captures the shadow of the wrinkle and local geometry, while inter-part occlusion (calculated when Gaussian was transformed to the canonical space of other parts) models \eg, shadow cast by the body onto the inner side of the arm.  Posed body geometry can be used with Monte Carlo (MC) ray-tracing to compute shadows but it's not real-time.  Our pre-computed part-wise occlusion probes avoid MC ray-tracing, enabling real-time rendering. 

\subsection{Worse Albedo but Better Relighting Results} 
The ground-truth dataset and the teacher model both employ MC ray tracing.  If we use a large loss to enforce the albedo from the teacher to the student, the final relighting results will be suboptimal since split-sum is an approximation to ray-tracing.  Hence, we use a small weight for albedo distillation, which serves more like a regularization to make sure the student does not produce unreasonable albedos.  Our albedos are thus optimized for split-sum and are not consistent with the ground-truth, which employs ray-tracing.
On the other hand, our learned normals are less noisy than the teacher, while split-sum does not suffer from MC noise.  These factors combined give us a better relighting result.

\section{Limitations and Societal Impact Discussion}
The final quality of our approach largely depends on the stability of the
teacher model.  Currently, the teacher model~\cite{Wang2024CVPR} requires
accurate body pose estimation and foreground segmentation, which may not be the
case for in-the-wild captures.  Combining existing state-of-the-art in-the-wild
avatar models~\cite{Chen2023CVPR,Liu2023ICCV,Jiang2024CVPR} with our efficient
relightable model is an interesting direction for future work.

Furthermore, the ambient occlusion assumption in our method may not hold in
the presence of strong point lights.  In such cases, the shading model may not be
able to capture the correct shadowing effects.  Also, similar to other
state-of-the-art models~\cite{Xu2024CVPR,Wang2024CVPR,Lin2024AAAI}, our model
can only handle direct illumination at inference time.  Modeling global
illumination effects while still achieving real-time performance is an active
area of research in both computer graphics and computer vision.

Moreover, our current per-scene optimization-based pipeline remains slow during training. Similar to other state-of-the-art feed-forward dynamic reconstruction methods~\cite{jiang2025geo4d, stream3r2025, wang2025pi3}, a promising future direction is to learn a data-driven prior for intrinsic property decomposition, enabling a feed-forward approach for animatable and relightable avatar reconstruction.

Regarding the societal impact, our work can be used to create realistic
avatars for virtual reality, gaming, and social media.  However, it is important
to consider the ethical implications of using such technology.  For example, our
method can be used to create deepfakes, which can be used to spread
misinformation.  It is important to develop methods to detect deepfakes and
educate the public about the existence of such technology.

\tablesupprana

\end{document}